\def\year{2022}\relax
\documentclass[letterpaper]{article} % DO NOT CHANGE THIS
\usepackage{aaai22}  % DO NOT CHANGE THIS
\usepackage{times}  % DO NOT CHANGE THIS
\usepackage{helvet}  % DO NOT CHANGE THIS
\usepackage{courier}  % DO NOT CHANGE THIS
\usepackage[hyphens]{url}  % DO NOT CHANGE THIS
\usepackage{graphicx} % DO NOT CHANGE THIS
\usepackage{natbib}  % DO NOT CHANGE THIS AND DO NOT ADD ANY OPTIONS TO IT
\usepackage{caption} % DO NOT CHANGE THIS AND DO NOT ADD ANY OPTIONS TO IT
\DeclareCaptionStyle{ruled}{labelfont=normalfont,labelsep=colon,strut=off} % DO NOT CHANGE THIS
\usepackage{algorithm}
\usepackage{algorithmic}

\usepackage[utf8]{inputenc}
\usepackage{listings}
\usepackage{xcolor}
\usepackage{booktabs}

\usepackage{pythontex} 

\newcommand{\mytilde}{\raise.17ex\hbox{$\scriptstyle\mathtt{\sim}$}}

\definecolor{dkgreen}{rgb}{0,0.6,0}
\definecolor{gray}{rgb}{0.5,0.5,0.5}
\definecolor{mauve}{rgb}{0.58,0,0.82}

\title{
Dichotomic Pattern Mining with Applications to \\
Intent Prediction from Semi-Structured Clickstream Datasets
}
\author{
    Xin Wang
    and 
    Serdar Kad{\i}o\u{g}lu
    \\
}
\affiliations{
    AI Center of Excellence, Fidelity Investments, Boston, USA\\
    \{firsname.lastname\}@fmr.com
}

\usepackage{amsthm}
\usepackage{amsfonts}
\theoremstyle{definition}

\usepackage{xcolor}

\usepackage{textcomp}

\usepackage{bibentry}
\begin{document}

\maketitle

\begin{abstract}
We introduce a pattern mining framework that operates on semi-structured datasets and exploits the dichotomy between outcomes. Our approach takes advantage of constraint reasoning to find sequential patterns that occur frequently and exhibit desired properties. This allows the creation of novel pattern embeddings that are useful for knowledge extraction and predictive modeling. Finally, we present an application on customer intent prediction from digital clickstream data. Overall, we show that pattern embeddings play an integrator role between semi-structured data and machine learning models, improve the performance of the downstream task and retain interpretability. 

\end{abstract}

% \vspace{-0.3cm}
\section{Introduction}
Intent prediction is an integral part of designing digital experiences that are geared toward user needs. Successful applications of intent prediction boost the performance of machine learning algorithms in various domains, including recommendation systems in e-commerce, virtual agents in retail, and conversational AI in the enterprise.

Intent prediction is a specific learning task as part of Knowledge Discovery. Knowledge extraction from various data sources has gained attraction in recent years as we become digitally connected more than ever before. 
% In addition, we witnessed an unprecedented increase in the number of AI-powered applications, which in return, generates even more data to harness. 
Recent work in this area expanded our capabilities from processing structured data such as traditional databases to unstructured data such as text, image, and video. 

In this paper, we consider digital clickstream as a source of \textit{semi-structured data}. On the one hand, the clickstream data provides \textit{unstructured text}, such as web pages. On the other hand, it yields sequential information where visits can be viewed as \textit{structured event streams} representing customer journeys. Given the clickstream behavior of a set of users, we are interested in two specific questions ranging from population-level to individual-level information extraction. At the population level, we are interested in finding the most frequent clickstream patterns across all users subject to a set of properties of interest. At the individual level, we are interested in downstream tasks such as intent prediction. Finally, an overarching theme over both levels is the interpretability of the results. 

Our contributions show that i) constrained-based sequential pattern mining is effective in extracting knowledge from semi-structured data and ii) serves as an integration technology to enable downstream applications while retaining interpretability. The main idea behind our approach is first to capture the population characteristics and extract succinct representations from large volumes of digital clickstream activity. Then, the patterns found become consumable in machine learning models. Overall, our generic framework alleviates manual feature engineering, automates feature generation, and improves the performance of downstream tasks.

To demonstrate our approach, we explore a public clickstream dataset with positive and negative intents for product purchases based on shopping activity. We apply our framework to find the most frequent patterns in digital activity and then leverage them in machine learning models for intent prediction. Finally, we show how to extract high-level signals from patterns of interest. 

% Knowledge discovery and extraction from various data sources have gained attraction in the recent years with the unprecedented increase in the number of AI-powered applications serving users and as we have become digitally connected more than ever before. Intent prediction is one such machine learning task for knowledge extraction. Intent prediction based on the digital interactions of users remains a challenging tasks because it relies on our ability to reason about vast amounts of unstructured, such as text, and semi-structured data, such as digital clickstream. 

% As we are moving toward the era where more customers are getting digitally connected, how we proactively engage with the customers through digital marketing becomes crucial for the thriving of the business. Customer intent prediction is the technique that has been used to boost the digital experience by anticipating customer needs in advance based on the previous interactions. It has been applied in many domains such as Recommender Systems in E-commerce, Conversational AI solutions and financial services in retail banking. 
% \todo{citations}

\begin{table}[t]
\centering
\renewcommand{\arraystretch}{1.3}
\begin{tabular}{|c|}
\hline
\textsc{Sequence Database} $\langle$(item, price, timestamp)$\rangle$                         \\ \hline
\hline
$\langle$ (A, 5, 1), (A, 5, 1), (B, 3, 2), (A, 8, 3), (D, 2, 3)$\rangle$ \\ \hline
$\langle$(C, 1, 3), (B, 3, 8), (A, 3, 9)$\rangle$                        \\ \hline
$\langle$(C, 4, 2), (A, 5, 5), (C, 2, 5), (D, 1, 7)$\rangle$           \\ \hline
\end{tabular}
\vspace{-0.1cm}
\caption{Example sequence database with three sequences.}
\vspace{-0.4cm}
\label{tab:example}
\end{table}

\section{Mining Clickstream Datasets}
Sequential Pattern Mining (SPM) is relevant for customer intent prediction, especially for sequential digital activity. Applications of SPM include 
% the analysis of medical treatment history~\cite{MedicalDataMining}, 
customer purchases, call patterns and digital clickstream~\citep{Requena2020}.
% ~\cite{Agrawal:Mining:1995, Srikant:Mining:1996}.
% A recent survey can be found in~\cite{Gan:Survey:2019}.

In SPM, we are given a set of sequences that is referred to as \textit{sequence database}. As shown in the example in Table~\ref{tab:example}, each sequence is an ordered set of \textit{items}. Each item might be associated with a set of \textit{attributes} to capture item properties, e.g., price, timestamp. A \textit{pattern} is a subsequence that occurs in at least one sequence in the database maintaining the original ordering of items. The number of sequences that contain a pattern defines the \textit{frequency}. Given a sequence database, SPM aims to find patterns that occur more than a certain frequency threshold. Here, we find three frequent patterns: [A, D] in rows 1 and 3, [B, A] in rows 1 and 2 and [C, A] in rows 2 and 3, each occurring in two sequences. 

% As a result,we  discover  three  patterns{[A,D],[B,A],[C,A]}subjectto minimum frequency threshold. Notice that each patternoccurs in exactly two sequences satisfying the minimum fre-quency. More specifically;•  The pattern[A,D]is a subsequence of the first and thethird sequence.•  The pattern[B,A]is a subsequence of the first and thesecond sequence.•  The pattern[C,A]is a subsequence of the second and thethird sequence.

In practice, finding the entire set of frequent patterns in a sequence database is not the ultimate goal. The number of patterns is typically too large and may not provide significant insights. It is thus important to search for patterns that are not only frequent but also capture specific properties of the application. This has motivated research in Constraint-based SPM (CSPM)
% ~\citep{CSPM_growth, Chen:Efficient:2008}
~\citep{Chen:Efficient:2008}. The goal of CSPM is to incorporate constraint reasoning into sequential pattern mining to find smaller subsets of interesting patterns. 

As an example, let us consider online retail clickstream analysis. We might not be interested in all frequent browsing patterns. For instance, the pattern $\langle login, logout\rangle$ is likely to be frequent but offers little value. Instead, we seek recurring clickstream patterns with unique properties, e.g., frequent patterns from sessions where users spend at least a minimum amount of time on a particular set of items with a specific price range. Such constraints help reduce the search space for the mining task and help discover patterns that are more effective in knowledge discovery than arbitrary patterns. 

\begin{algorithm}[t]
\caption{Dichotomic Pattern Mining}
\label{algo}
\begin{algorithmic} 
    \STATE{\textbf{In:}} \textit{Sequence database} $\mathcal{SD}$ 
    % and attributes $\mathbb{A}$
    % \STATE{\textbf{In:}} \textit{Attribute database} $\mathbb{A}$
    \STATE{\textbf{In:}} \textit{Binary label for sequences} $\mathcal{Y}$
    \STATE{\textbf{In:}} \textit{Minimum frequency threshold} $\mathcal{\theta}$
    \STATE{\textbf{In:}} \textit{Pattern constraints} $C_{type}(\cdot)$
    \STATE{\textbf{Out:}} \textit{Frequent pattern sets $\mathcal{P}$}
    % Patterns for downstream analysis or modeling tasks} 
    % \STATE{\textbf{Out:}} \textit{Trained Machine Learning model for intent prediction} $\mathcal{M}$

    \item[]
    \STATE{\textbf{Step 1.}} Dichotomic split over the dataset
    \STATE $Pos \leftarrow \{SD_i \mid Y_i = \top \} $
    \STATE $Neg \leftarrow \{SD_i \mid Y_i = \bot \} $
    
    \STATE 
    \STATE{\textbf{Step 2.}} Apply constraint-based frequent pattern mining
    \STATE $Pos_{frequent} \leftarrow CSPM(Pos, ~ C_{type}(\cdot), ~ \mathcal{L}) $
    \STATE $Neg_{frequent} \leftarrow CSPM(Neg, ~ C_{type}(\cdot), ~ \mathcal{L})$
    
    \STATE 
    \STATE{\textbf{Step 3.}} Find unique patterns and their union 
    \STATE $Pos_{unique} \leftarrow Pos_{frequent} \setminus Neg_{frequent}$
    \STATE $Neg_{unique} \leftarrow Neg_{frequent} \setminus Pos_{frequent}$
    \STATE $PN_{union} \leftarrow Pos_{frequent} \cap Neg_{frequent}$
    
    \STATE 
    \STATE{\textbf{Step 4.}} Return frequent patterns for downstream tasks
     \STATE $\mathcal{P} \leftarrow  \{Pos_{unique},  Neg_{unique}, PN_{union}\}$
    \STATE return $\mathcal{P}$ 
    
    % \STATE{\textbf{1.}} ~~Split data $\mathcal{SD}$ and $\mathbb{A}$ into two partitions according to the binary labels $\mathcal{Y}$.
    
    % \STATE{\textbf{2.}} ~~Apply CSPM separately on the positive and negative data partitions to identify patterns $\mathcal{P}_{positive}$ and $\mathcal{P}_{negative}$, satisfying the constraints $C_{type}(\cdot)$ and minimum pattern frequency $\mathcal{L}$.
    
    % \STATE{\textbf{3.}} ~~Compare $\mathcal{P}_{positive}$ and $\mathcal{P}_{negative}$, to find unique patterns in either one of the two sets, $\mathcal{P}_{positive\_unique}$ and $\mathcal{P}_{negative\_unique}$, and also find the patterns in common $\mathcal{P}_{common}$.
    
    % \STATE{\textbf{4.}} ~~Finalize the patterns for downstream tasks, $\mathcal{P}$, utilizing the patterns found in step 3.
    
    %  \STATE{\textbf{5.}} ~~Each sequence in $\mathcal{SD}$ is represented by a feature vector based on an encoding of the existence of individual patterns in $\mathcal{P}$.
     
    %  \STATE{\textbf{6.}} ~~Train model $\mathcal{M}$ using the extracted data features and labels $\mathcal{Y}$.

\end{algorithmic}
\end{algorithm}

\section{Dichotomic Pattern Mining}
We now describe our dichotomic pattern mining approach that operates over sequence databases augmented with binary labels denoting positive and negative outcomes. In our application, we use intent prediction as the outcome.
% , intent in this paper. The algorithm enables feature extraction, which are then later for intent prediction. 

% In this section, we demonstrate how to automate the feature extraction process leveraging CSPM to identify patterns, then the generated features will serve as the input for downstream machine learning models to predict customer intent. 

Algorithm \ref{algo} presents our generic approach. The algorithm receives a sequence database, $\mathcal{SD}$, containing \textit{N} sequences $\{S_1, S_2, \ldots, S_N\}$. Each sequence represents a customer's behaviors in time order, for example, the digital clicks in one session. Sequences are associated with binary labels, $\mathcal{Y}$, indicating the outcome of the sequence to be positive or negative, e.g., purchase or non-purchase. As in our example in Table~\ref{tab:example}, the items in each sequence are associated with a set of attributes $\mathbb{A} = \{\mathcal{A}, \ldots, \mathcal{A_{|\mathbb{A}|}}\}$.  There is a set of functions $C_{type}(\cdot)$ imposed on attributes with a certain type of operation. For example, $C_{avg}(\mathcal{A}_{price})\ge 20$ requires a pattern to have minimum average price 20. Similarly, there is a minimum threshold $\mathcal{\theta}$ as frequency lower bound. 

% $\mathcal{P}$ is the set of identified frequent patterns for knowledge discovery and downstream Machine Learning models, $\mathcal{M}$ is your Machine Learning model that can be built.

Our algorithm is conceptually straightforward and exploits the dichotomy between outcomes. At a high level, we first split the sequences into positive and negative sets. We then apply CSPM on each group separately subject to minimum frequency while satisfying pattern constraints. Notice that frequent patterns found might overlap. Therefore, we perform a set difference operation in each direction. This allows us to distinguish between recurring patterns that \textit{uniquely} identify the positive and negative populations. 

The output of Algorithm~\ref{algo} is a set of frequent patterns, $\mathcal{P}$, that provides insights into how the sequential behavior varies between populations. Thus, our algorithm serves as an integration block between pattern mining algorithms, CSPM in this paper , and the learning task, intent prediction in this paper. Using $\mathcal{P}$, we learn new representations for sequences. To create a feature vector for each sequence, we encode them using patterns. A typical approach is one-hot encoding to indicate the existence of patterns. Overall, this approach yields an automated feature extraction process that is generic and independent of the subsequent machine learning models applied to pattern embeddings.

% In Algorithm \ref{algo}, we show that the CSPM can be applied separately to the data from positive intent or negative intent sequences. With a comparison of these two sets of patterns, we get insights for how the sequential behaviors may vary between the two populations. To create a feature vector for each sequence, we finalize the patterns of interests and have each sequence to be encoded using the patterns. A typical approach is via one-hot encoding to indicate the existence of patterns. The automated feature extraction process is generic and independent of the subsequent models, thus various supervised machine learning models can be trained using the created data features and labels. 

% In the inference stage, a sequence of behaviors will be encoded based on the interested patterns, following the same feature extraction procedure in Algorithm \ref{algo} at Step 5. Then the trained model is applied to the data for making intent predictions. 

% In the next section, we present a demonstration of an experiment by applying the approach on a public data set.

% for creating the features the auto-generated features by \texttt{Seq2Pat} match the hand-crafted features explicitly designed for this dataset and improves the performance of predictive models.

\begin{table}[t]
\renewcommand{\arraystretch}{1.3}
\centering
\begin{tabular}{|c|c|}
\hline
\textsc{Symbol} &  \textsc{Event}\\ 
\hline
\hline
1 & Page view \\ \hline
2 & Detail (see product page)\\ \hline
3 & Add (add product to cart)\\ \hline
4 & Remove (remove product from cart) \\ \hline
5 & Purchase\\ \hline
6 & Click (click on result after search)\\
\hline
\end{tabular}%
\caption{The symbols used to depict clickstream events.}
\label{tab:symbols}
\vspace{-0.5cm}
\end{table}

\vspace{-0.05cm}
\section{Customer Intent Prediction}
We apply our algorithm for customer intent prediction from click sequences of online shoppers. In the following, we describe the data, the constraint-based pattern mining, feature generation, and prediction models. We then present numeric results and study feature importance to drive insights and explanations from auto-generated features.

\subsection{Clickstream Dataset} 
The dataset contains rich clickstream behavior on online users browsing a popular fashion e-commerce website~\citep{Requena2020}. It consists of 203,084 shoppers' click sequences. There are 8,329 sequences with at least one purchase, while 194,755 sequences lead to no purchase. The sequences are composed of symbolized events as shown in Table~\ref{tab:symbols} with length $L$ between the range $5 \le L \le 155$. 
% For sequences having a purchase, we only keep the events preceding the first purchase such that the purchase symbols are removed from the data. 
Sequences leading to purchase are labeled as positive (+1); otherwise, labeled as negative (0), resulting in a binary intent classification problem.

\begin{table*}[t]
\renewcommand{\arraystretch}{1.3}
\centering
\begin{tabular}{|c|c|c|c|c|c|}
\hline
\textsc{Model} & \textsc{Features Space} & \textsc{Precision(\%)} & \textsc{Recall(\%)} & \textsc{F1(\%)} & \textsc{AUC(\%)}\\ 
\hline
\hline
\texttt{LightGBM} & Seq2Pat Patterns & 44.70 ($\pm$~1.92) & 63.15 ($\pm$~4.65) & 52.20 ($\pm$~0.65) & 94.98 ($\pm$~0.15)\\ \hline
\texttt{Shallow\_NN} & Seq2Pat Patterns & 44.40 ($\pm$~2.18) & 64.11 ($\pm$~4.57) & 52.31 ($\pm$~0.54)& 95.00 ($\pm$~0.17)\\ \hline
\texttt{LSTM} & Clickstream  & \textbf{54.96} ($\pm$~1.77) & 69.53 ($\pm$~4.31) & 61.28 ($\pm$~0.95) & 96.41 ($\pm$~0.15)\\ \hline
\texttt{LSTM\_Seq2Pat} & Clickstream + Seq2Pat Patterns & 54.35 ($\pm$~2.40) & \textbf{73.64} ($\pm$~4.70) & \textbf{62.39} ($\pm$~0.81) & \textbf{96.76} ($\pm$~0.12)\\ \hline
\end{tabular}%
\vspace{-0.1cm}
\caption{Comparison of averaged intent classification performance by different methods over 10 random Train-Test splits.}
\label{tab:performance}
\vspace{-0.4cm}
\end{table*}

\subsection{Constrained-based SPM}\label{apply_cspm}
In Algorithm~\ref{algo} at Step 2, we need a data mining approach to extract frequent patterns. For that purpose, we utilize \texttt{Seq2Pat}~\cite{seq2pat} to find sequential patterns that occur frequently. \texttt{Seq2Pat} supports constraint-based reasoning to specify desired properties. It uses the state-of-the-art multi-valued decision diagram representation of sequences~\citep{HosseininasabHC19}. 

Next we declare our constraint model, $C_{type}(\cdot)$, to specify patterns of interest. For each event, we have two attributes: the sequential order in a sequence, $\mathcal{A}_{order}$, and the dwell time on a page, $\mathcal{A}_{time}$. 
We enforce two constraints to seek interesting patterns. First, we require the maximum length of a pattern to be 10. Additionally, we seek page views where customers spend at least 20 secs on average. More precisely, we set $C_{span}(\mathcal{A}_{order})\le 10$ and $C_{avg}(\mathcal{A}_{time})\ge 20_{(sec)}$. We set the minimum frequency threshold  $\mathcal{\theta}$ as the 30$\%$ of the total number of sequences. 

% We divide the shoppers into two groups as purchasers and non-purchasers. Then, we apply \texttt{Seq2Pat} to each group independently to mine patterns specific to each characteristic.

% We set the maximal length of the sequences to be 20, such that longer sequences are trimmed to contain only the last 20 events. This effectively reduces computational costs and speeds up the mining process. 
% Note that the maximum span constraint is anti-monotone and is relatively easy to enforce. In contrast, the average constraint is non-monotone and is one of the challenging constraints that is supported uniquely by \texttt{Seq2Pat}. 
% The original data set has sequences subject to a time constraint by limiting the sequences to be within 30 minutes as one session, such that we require no more restrictions on the total time span of patterns. 
With this constraint model, \texttt{Seq2Pat} finds 457 frequent patterns in purchase sequences,  $Pos_{frequent}$, and 236 frequent patterns from the non-purchase sequences, $Neg_{frequent}$, with some overlap between the two groups.

% In our works, we have been utilizing \texttt{Seq2Pat} for implementing CSPM. \texttt{Seq2Pat} is a tool built in collaboration between academia and industry to serve researchers and practitioners for knowledge discovery in large sequence databases. The tool finds sequential patterns that occur frequently. Furthermore, it supports constraint-based reasoning to specify desired properties by leveraging the state-of-the-art multi-valued decision diagram representation of the sequence database~\citep{HosseininasabHC19}. One unique advantage of \texttt{Seq2Pat} is the support of several complex non-monotone constraint types, such as average or median constraints.

\vspace{-0.05cm}
\subsection{Feature Generation}
\label{sec:features}
When the sets of patterns from purchaser and non-purchaser are compared, we find 244 unique purchaser patterns, $Pos_{unique}$, and 23 unique non-purchaser patterns, $Neg_{unique}$. The groups share 213 patterns in common. In combination, we have 480 unique patterns $PN_{union}$. We generate the feature space via one-hot encoding. For each sequence, we create a 480-dimensional feature vector with a binary indicator to denote the existence of a pattern.
% by using the union of the two sets of patterns that is composed of 480 patterns. 
% Then each sequence is represented by a 480 dimensional feature vector with each feature is a binary indicator of a pattern being detected. 

\vspace{-0.05cm}
\subsection{Intent Modeling} 
To study the the behaviour of auto-generated features we develop four different models to predict customer intent: 
\begin{enumerate}
    \item \texttt{LightGBM}
    % ~\citep{Ke:lightgbm:2017} 
    over the \texttt{Seq2Pat} patterns.  
    \item \texttt{Shallow\_NN} shallow neural network using one hidden layer over the \texttt{Seq2Pat} patterns.
 \item \texttt{LSTM} Long short-term memory network
%  ~\citep{Sepp:LSTM:1997} 
 from~\citep{Requena2020} that uses input sequences as-is. \texttt{LSTM} applies one hidden layer on the output of the last layer followed by a fully connected layer to make intent prediction.
 \item \texttt{LSTM\_Seq2Pat} The \texttt{LSTM} model boosted with pattern embeddings. \texttt{LSTM\_Seq2Pat} uses the same architecture with \texttt{LSTM}, the only difference being \texttt{Seq2Pat} based features are concatenated to the output of \texttt{LSTM} and are used together as input of the hidden layer. 
\end{enumerate} 

Notice that simpler models such as \texttt{LightGBM} and \texttt{Shallow\_NN} cannot operate on semi-structured clickstream data since they cannot accommodate recurrent sequential relationships. Contrarily, more sophisticated architectures, such as \texttt{LSTM} can work directly with the input. For the former, our approach allows simple models to work with sequence data. For the latter, our approach augments advanced models by incorporating pattern embeddings into the feature space.  

\subsection{Model Training} 

We use 80$\%$ of the data as the train set and 20$\%$ as the test set and repeat this split 10 times for robustness. We compare the average results for each model based on Precision, Recall, F1 score, and the area under the ROC curve, aka AUC.  

\smallskip
\noindent{\textbf{Hyper-parameter Tuning:}} We apply 3-fold cross-validation for hyper-parameter tuning in the first train-test split. We apply grid search on the number of iterations [400, 600, 800, 1000] for \texttt{LightGBM}, number of nodes in the hidden layer [32, 64, 128, 256, 512] for \texttt{Shallow\_NN} and \texttt{LSTM} models, number of \texttt{LSTM} units [32, 64, 128]. We use 10$\%$ of train set as a validation set to determine if training meets early stop condition. When the loss on validation set stops decreasing steadily, training is terminated. The validation set is used to determine a decision boundary on the predictions for the highest F1 score. The final parameters are 400 iterations for \texttt{LightGBM}, 64 nodes for \texttt{shallow\_NN}, 32 \texttt{LSTM} units in \texttt{LSTM} models with 64 and 128 nodes in hidden layers.

\subsection{Prediction Performance} 
Table~\ref{tab:performance} presents the average results that compare the performance of the models. For feature space, we either use the patterns found by \texttt{Seq2Pat}, the original clickstream events in the raw data, or their combination. Using auto-generated \texttt{Seq2Pat} features, \texttt{LightGBM} and \texttt{Shallow\_NN} models achieve a performance that closely match the results given in the reference work~\citep{Requena2020}. The difference is, models in~\citep{Requena2020} use hand-crafted features, while we automate the feature generation process here. When a more sophisticated model such as \texttt{LSTM} is used, it outperforms \texttt{LightGBM} and \texttt{Shallow\_NN}. When the \texttt{LSTM} model is combined with \texttt{Seq2Pat}, \texttt{LSTM\_Seq2Pat} yields  substantial increase in Recall, and consequently, the highest F1 score. \texttt{LSTM\_Seq2Pat} is also superior to others in terms of AUC. We conclude that the features extracted automatically via \texttt{Seq2Pat} boost ML models in the downstream task for shopper intent prediction.

% \begin{figure}[t]
% \centering
% \includegraphics[width=0.75\columnwidth]{time.png}
% \caption{The rune time and number of mined patterns as we increase the minimum time span constraint on the clickstream sequences having a purchase.
% % \textit{min\_frequency}=2.
% }
% \label{fig:time}
% \end{figure} 

\begin{figure}[t]
\centering
\includegraphics[width=0.75\columnwidth]{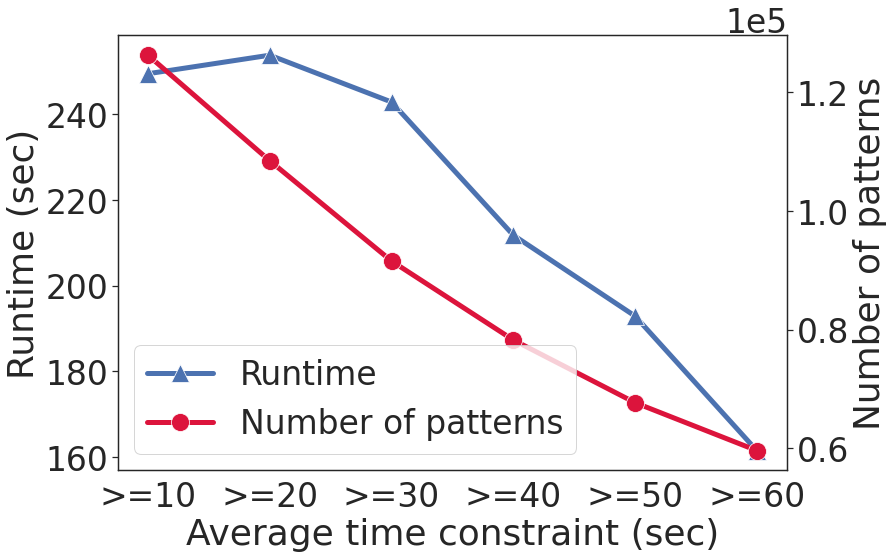}
\vspace{-0.1cm}
\caption{The runtime (y-axis-left) and the number of patterns found (y-axis-right) with varying constraints (x-axis). 
% and the minimum frequency threshold, $\theta$, set to 2, to stress test runtime performance. 
}
\label{fig:time}
\vspace{-0.5cm}
\end{figure} 

\subsection{Runtime Performance}
We report runtime performance of pattern mining on a machine with Linux RHEL7 OS, 16-core 2.2GHz CPU, and 64 GB of RAM. We apply \texttt{Seq2Pat} on the positive set with 8,329 clickstream sequences. We impose the same types of constraint as described in Section \ref{apply_cspm} while we vary the constraint on the minimum average time spent on pages.
To stress test the runtime, we set the minimum frequency $\theta=2$ which returns almost all the feasible patterns. 
Figure \ref{fig:time} shows the runtime in seconds (y-axis-left) and the number of patterns found (y-axis-right) as the average constraint increases (x-axis). As the constraint becomes harder to satisfy, the number of patterns goes down as expected. The runtime for the hardest case is \mytilde250 seconds while we observe speed-up as constraint reasoning becomes more effective. 

\vspace{-0.2cm}
\subsection{Feature Importance} 
Finally, we study feature importance to drive high-level insights and explanations from auto-generated \texttt{Seq2Pat} features. We examine the 
% Shapley Additive Explanation (SHAP)~\citep{shap_value} value
Shapley value~\citep{shap_value}
of features from the \texttt{LightGBM} model. 

\begin{figure}[t]
\includegraphics[width=\columnwidth]{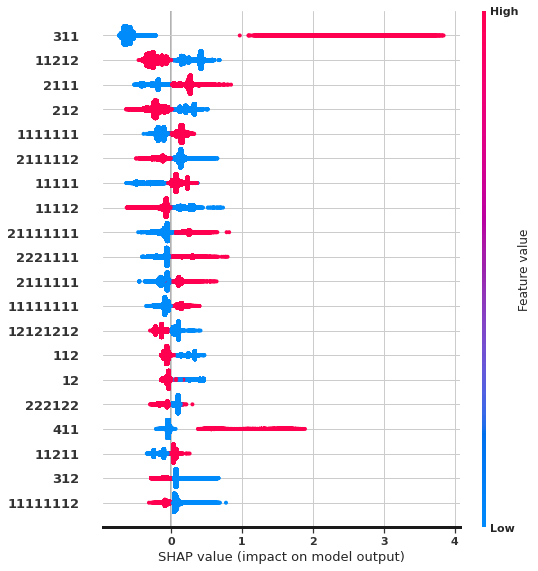}
\caption{SHAP values of auto-generated \texttt{Seq2Pat} features. Top-20 features  ranked in descending importance. Color indicates high (in red) or low (in blue) feature value. Horizontal location indicates the correlation of the feature value to a high or low model prediction.}
\label{fig:shap}
\vspace{-0.4cm}
\end{figure}

Figure~\ref{fig:shap} shows the top-20 features with highest impact. Our observations match  previous findings in~\citep{Requena2020}. The pattern $\langle3, 1, 1\rangle$ provides the most predictive information, given that the symbol (3) stands for adding a product. Repeated page views as in $\langle1,1,1,1,1,1,1\rangle$, or specific product views, $\langle2,1,1,1\rangle$ are indicative of purchase intent, whereas web exploration visiting many products, $\langle1,1,2,1,2\rangle$, are more negatively correlated to a purchase. Interestingly, searching actions $\langle6\rangle$ have minimum impact on buying, raising questions about the quality of the search and ranking systems. Our frequent patterns also yield new insights not covered in the existing hand-crafted analysis. Most notably, we discover that removing a product but then remaining in the session for more views, $\langle4,1,1\rangle$ is an important feature, positively correlated with a purchase. This scenario, where customers have specific product needs, hints at missed business opportunity to create incentives such as prompting virtual chat or personalized promotions.

% \section{Related Work}
% Historically, SPM was introduced in the context of market basket analysis~\cite{Agrawal:Mining:1995} with several algorithm such as GSP~\cite{Srikant:Mining:1996}, PrefixSpan~\cite{jian:prefixspan:2001}, SPADE~\cite{zaki:spade:2001} and SPAM~\cite{ayres:sequential:2002}. Mining the complete set of patterns imposes high computational costs and contains a large number of redundant patterns. Thus CSPM is proposed to alleviate this problem~\cite{bonchi:pushing:2005, Nijssen2014, aoga:mining:2017}. Constraint Programming and graphical representation of the sequence database have been shown to perform well for CSPM~\cite{kemmar:prefix:2017, GUNS:miningzinc:2017, borah:FP:2018, HosseininasabHC19}.

% Although a few Python libraries exist for SPM, see, e.g., \cite{PrefixSpan-py, pymining}, to the best of our knowledge, \texttt{Seq2Pat} is the first CSPM library in Python that supports several anti-monotone and non-monotone constraint types. Unfortunately, other CSPM implementations are either not available in Python, hence missing the opportunity to integrate with ML applications, or limited to a few constraint types, most commonly, gap, maximum span, and regular expressions~\citep{Yu:Generalized:2006, ccspan, aoga2016efficient, Fouriner:SPMF:2016}.

\vspace{-0.2cm}
\section{Conclusion}
Pattern mining is an essential part of data analytics and knowledge discovery from sequential databases. It is a powerful tool, especially when combined with constraint reasoning to specify desired properties. In this paper, we presented a simple procedure for Dichotomic Pattern Mining that operates over semi-structured clickstream datasets. The approach learns new representations of pattern embeddings. This representation enables simple models, which cannot handle sequential data by default, to predict from sequences. Moreover, it boosts the performance of more complex models with feature augmentation. Experiments on customer intent prediction from fashion e-commerce demonstrate that our approach is an effective integrator between automated feature generation and downstream tasks. Finally, as shown in our feature importance analysis, the representations we learn from pattern embeddings remain interpretable.
% We hope our approach help improve improve applied AI innovation and deployment

\footnotesize
% \scriptsize
% \vspace{-0.2cm}
\bibliography{main.bib}

\end{document}